\documentclass[runningheads]{llncs}
\usepackage{graphicx}
\usepackage{amsmath}
\usepackage{multirow}
\usepackage{booktabs}
\usepackage{array}
\usepackage{subfigure}
\usepackage{amssymb}
\usepackage{algorithm}
\usepackage{algorithmic}

\begin{document}

\title{FAQ-based Question Answering via Knowledge Anchors}

\author{Ruobing Xie\thanks{Corresponding author: Ruobing Xie} \and Yanan Lu \and Fen Lin \and Leyu Lin}

\institute{WeChat Search Application Department, Tencent, China\\
\email{ruobingxie@tencent.com}}

\maketitle              

\begin{abstract}
Question answering (QA) aims to understand questions and find appropriate answers. In real-world QA systems, Frequently Asked Question (FAQ) based QA is usually a practical and effective solution, especially for some complicated questions (e.g., How and Why). Recent years have witnessed the great successes of knowledge graphs (KGs) in KBQA systems, while there are still few works focusing on making full use of KGs in FAQ-based QA. In this paper, we propose a novel Knowledge Anchor based Question Answering (KAQA) framework for FAQ-based QA to better understand questions and retrieve more appropriate answers. More specifically, KAQA mainly consists of three modules: knowledge graph construction, query anchoring and query-document matching. We consider entities and triples of KGs in texts as knowledge anchors to precisely capture the core semantics, which brings in higher precision and better interpretability. The multi-channel matching strategy also enables most sentence matching models to be flexibly plugged in our KAQA framework to fit different real-world computation limitations. In experiments, we evaluate our models on both offline and online query-document matching tasks on a real-world FAQ-based QA system in WeChat Search, with detailed analysis, ablation tests and case studies. The significant improvements confirm the effectiveness and robustness of the KAQA framework in real-world FAQ-based QA.
\end{abstract}

\section{Introduction}

Question answering (QA) aims to find appropriate answers for user's questions. According to the type of answers, there are mainly two kinds of QA systems.
For simple questions like ``\emph{Who writes Hamlet?}", users tend to directly know the answers via several entities or a short sentence. KBQA is designed for these questions \cite{cui2016kbqa}. While for complicated questions like ``\emph{How to cook a risotto}?", users usually seek for detailed step-by-step instructions. In this case, \textbf{FAQ-based QA} system is a more effective and practical solution. It attempts to understand user questions and retrieve related documents as answers, which is more like a sentence matching task between questions and answers \cite{kothari2009sms}.

\begin{figure}[!htbp]
\centering
\includegraphics[width=0.85\columnwidth]{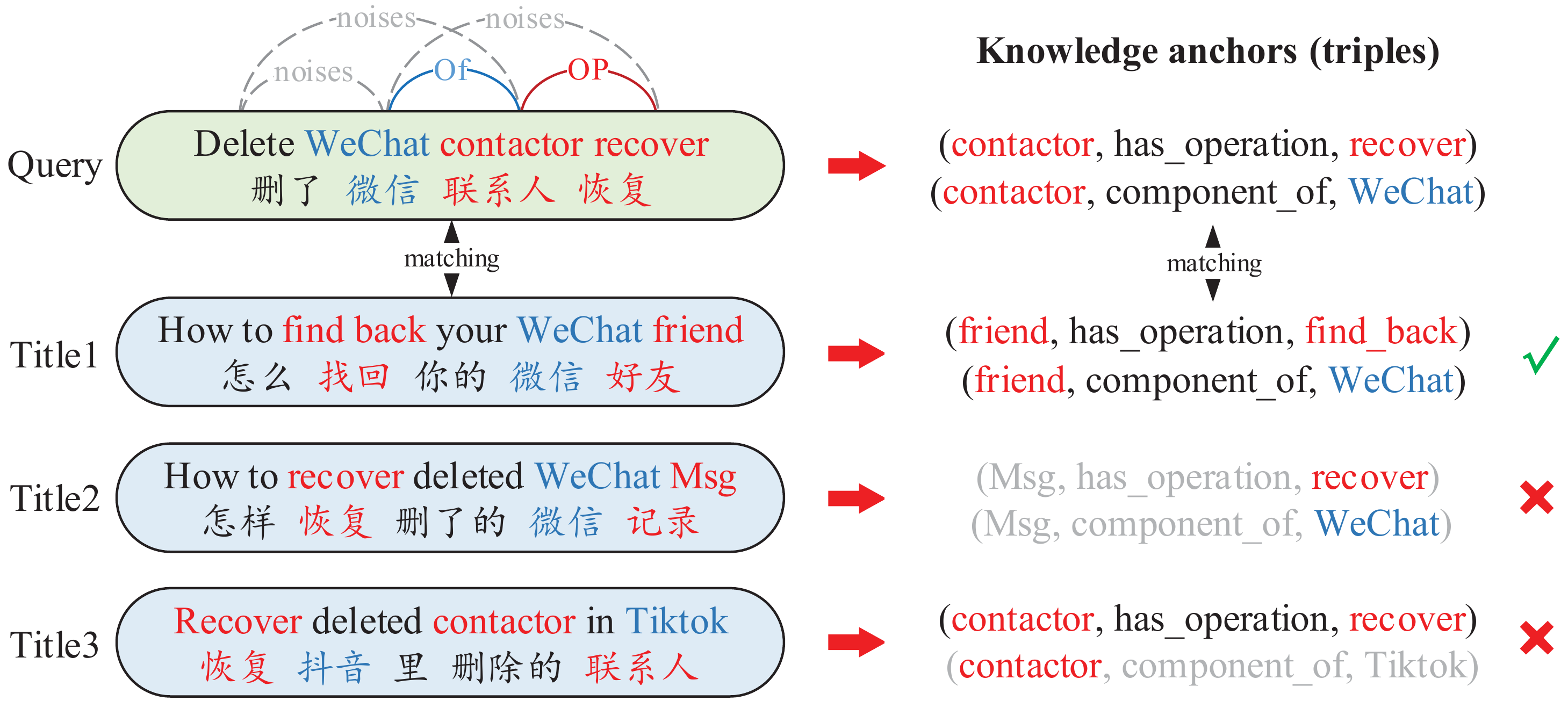}
\caption{An example of knowledge anchors (triples) in FAQ-based QA.}
\label{fig:example}
\end{figure}

QA systems always pursue higher precision and better interpretability, for users of QA systems are much more critical to the results compared to users in IR or dialog tasks. Recent years have witnessed the great thrive in knowledge graphs (KGs). A typical knowledge graph usually consists of entities, relations and triples, which can provide structural information for QA. KGs has been widely used in KBQA for simple questions \cite{cui2016kbqa}. However, there are still few works focusing on introducing KGs to FAQ-based QA for complicated questions.
The main challenge of FAQ-based QA is that its queries and answers are more difficult to understand, since complicated questions often involve with professional terms, domain-related operations and conditions.
A small semantic shift in the question may lead to a completely different answer. Moreover, the informal representations (e.g., VX), abbreviations (e.g., Msg) and domain-specific restrictions may further confuse the understanding and matching. Simply relying on conventional sentence matching models may not work well in this situation, dependency parsers and term weights trained on general corpus may also bring in errors.

To address these problems, we introduce KGs to FAQ-based QA systems. Differing from KBQA, we bring in KGs not to directly answer questions, but to \emph{better understand and match queries and titles}. A query/title in FAQ-based QA usually contains essential factors like \emph{entities} and \emph{triples} (i.e., entity pairs in texts with their relations) that derive from KGs. We consider such factors of KGs in query/title as \textbf{knowledge anchors}, which can anchor the core semantics in query and title for NLU and sentence matching.
Knowledge anchors can bring in \emph{higher precision} and \emph{better interpretability}, which also makes the FAQ system \emph{more robust and controllable}. Fig \ref{fig:example} gives an example of knowledge anchors in real-world queries and titles. The knowledge anchors bring in prior knowledge and highlight the core semantics as well as restrictions for matching.

In this paper, we propose a novel \textbf{Knowledge Anchor based Question Answering (KAQA)} framework for FAQ-based QA. Precisely, KAQA mainly consists of three modules:
(1) knowledge graph construction, which stores prior domain-specific knowledge.
(2) Query anchoring, which extracts core semantics in queries and documents with three triple disambiguation modules.
And (3) multi-channel query-document matching, which calculates the semantic similarities between queries and documents with token and knowledge anchor sequences.
The advantages of KAQA mainly locate in two points:
(1) KAQA is a simple and effective framework, which cooperates well with almost all sentence matching algorithms, and can be easily applied to other industrial domains.
(2) Knowledge anchors in KAQA make it possible to understand queries and titles accurately with fine-grained domain-specific knowledge. The structural interpretable KG also make the FAQ-based QA system more robust and controllable.

In experiments, we build a new dataset from a real-world Chinese FAQ-based QA system, and conduct both online and offline evaluations. The results show that knowledge anchors and KAQA are essential in NLU and sentence matching. We further conduct some analyses on query anchoring and knowledge anchors with detailed cases to better interpret KAQA's pros and cons as well as its effective mechanisms. The main contributions are concluded as follows:

\begin{itemize}
\item We propose a novel KAQA framework for real-world FAQ-based QA. The multi-channel matching strategy also enables models to cooperate well with both simple and sophisticated matching models for different real-world scenarios. To the best of our knowledge, KAQA is the first to explicitly utilize knowledge anchors for NLU and matching in FAQ-based QA.
\item We conduct sufficient online and offline experiments to evaluate KAQA with detailed analyses and cases. The significant improvements confirm the effectiveness and robustness of KAQA. Currently, KAQA has been deployed on a well-known FAQ system in WeChat Search affecting millions of users.
\end{itemize}

\section{Related Work}

\textbf{Question Answering.}
FAQ-based QA is practical and widely used for complicated questions. \cite{cavnar1994n} gives a classical n-gram based text categorization method for FAQ-based QA.
Since the performance of FAQ-based QA is strongly influenced by query-document matching, lots of efforts are focused on improving similarity calculations \cite{zhou2016learning}.
Knowledge graphs have been widely used in QA. Semantic parser \cite{kwiatkowski2013scaling}, information extraction \cite{yao2014information} and templates \cite{zheng2018question} are powerful tools to combine with KGs. Recently, Pre-train models and Transformer are also used for QA and reasoning \cite{clark2020transformers}.
\cite{zhang2018variational} focuses on multi-hop knowledge reasoning in KBQA, and \cite{huang2019knowledge} explores knowledge embeddings for simple QA. However, models in FAQ-based QA usually ignore or merely use entities as features for lexical weighting or matching \cite{bedue2018novel}.
To the best of our knowledge, KAQA is the first to use knowledge anchors for NLU and matching in FAQ-based QA.

\textbf{Sentence Matching.}
Measuring semantic similarities between questions and answers is essential in FAQ-based QA. Conventional methods usually rely on lexical similarity techniques \cite{robertson2009probabilistic}. Inspired by Siamese network, DSSM \cite{huang2013learning} and Arc-I \cite{hu2014convolutional} extract high-order features and then calculate similarities in semantic spaces. Arc-II \cite{hu2014convolutional} and MatchPyramid \cite{pang2016text} extract features from lexical interaction matrix. IWAN \cite{shen2017inter} explores the orthogonal decomposition strategy for matching. Pair2vec \cite{joshi2019pair2vec} further considers compositional word-pair embeddings. \cite{kim2019semantic} also considers recurrent and co-attentive information.
Our multi-channel model enables most of sentence matching models to be plugged in KAQA flexibly.

\section{Methodology}

We first give an introduction of the notations used in this paper. For a knowledge graph $\{E,R,T\}$, $E$ and $R$ represents the entity and relation set. We utilize $(e_h,r,e_t) \in T$ to represent a triple in KG, in which $e_h,e_t \in E$ are the head and tail entity, while $r \in R$ is the relation.
We consider the query $q$ and document $d$ as inputs, and simply use titles to represent the documents for online efficiency.
In KAQA, both queries and titles are labelled with knowledge anchors in query anchoring module. The \textbf{knowledge anchor} set in a query $A^q=\{E^q,T^q\}$ contains two sequences, namely the \emph{entity} sequence $E^q$ and the \emph{triple} sequence $T^q$, where entities and triples are arranged by their positions. The knowledge anchor set $A^d=\{E^d,T^d\}$ in document is the same as that in query.

\subsection{Overall Architecture}

\begin{figure*}[!htbp]
\centering
\includegraphics[width=0.89\textwidth]{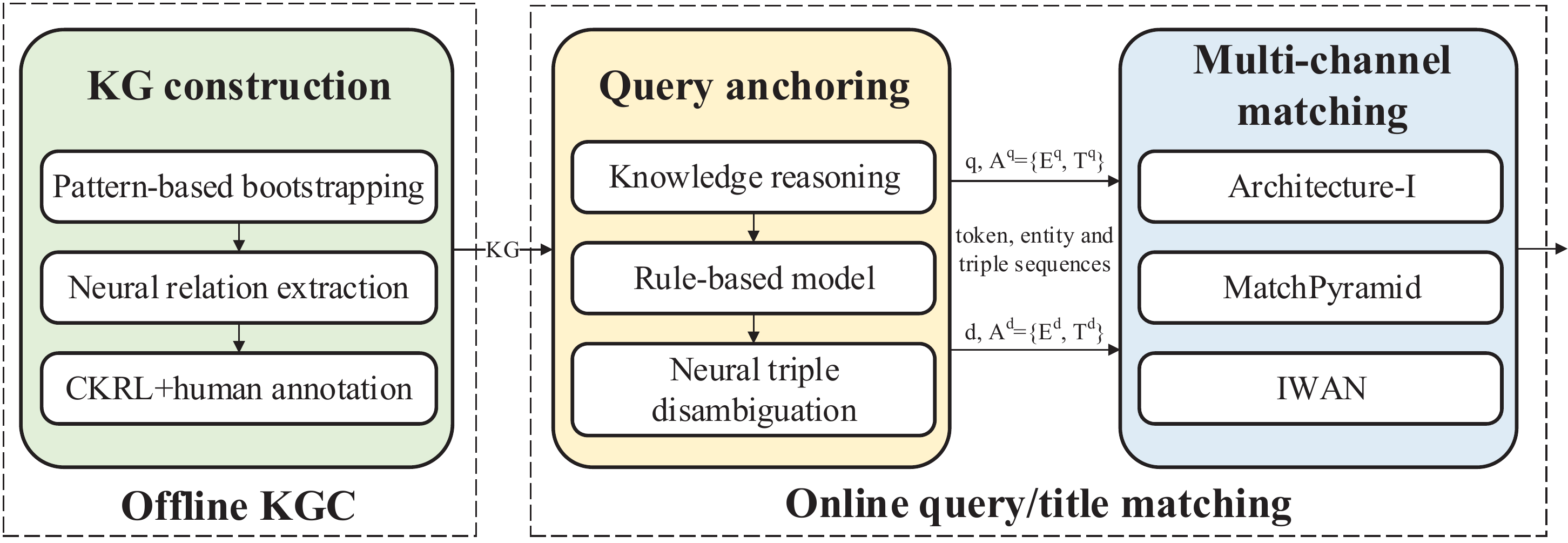}
\caption{The overall architecture of the KAQA framework.}
\label{fig:architecture}
\end{figure*}

KAQA mainly consists of three modules, namely knowledge graph construction, query anchoring and query-document matching. Fig. \ref{fig:architecture} shows the overall architecture of KAQA. Knowledge graph construction is the fundamental step to learn and store prior knowledge. Next, the query anchoring module scans queries and titles to extract knowledge anchors. Multiple disambiguation models are used to prove the reliability of extracted knowledge anchors. Finally, the query-document matching module measures the semantic similarity between queries and titles via their token, entity and triple sequences.

\subsection{Knowledge Graph Construction}

In KAQA, KGs are mainly utilized for better NLU and matching, not for directly answering questions. Therefore, instead of directly using existing open-domain KGs, we build a domain-specific customized KG, which focuses more on triples that represent core semantics in specific target domains. In most domains, the core semantics of a sentence are captured by triples like (\emph{contactor}, \emph{has\_operation}, \emph{recover}) as in Fig. \ref{fig:example}, which often imply certain actions on objects.
Specifically, we focus on the domain of software customer service. We mainly focus on four types of relations to anchor core semantics, namely \emph{has\_operation}, \emph{component\_of}, \emph{synonym} and \emph{hypernym/hyponym}. \emph{has\_operation} is responsible for the main operation, \emph{component\_of} reveals important relatedness, while \emph{synonym} and \emph{hypernym/hyponym} are used for alignment and generalization.

\begin{table}[!hbtp]
\centering
\small
\begin{tabular}{p{2.4cm}|p{8.4cm}}
\toprule
  query & How to \textbf{recover} WeChat \textbf{friend} if she has deleted me?  \\
\midrule
  entity & \emph{delete}; \emph{WeChat}; \emph{friend}; \emph{recover}; \\
\midrule
  triple & (\emph{WeChat}, \emph{has\_operation}, \emph{delete}) \\
  candidates & (\emph{WeChat}, \emph{has\_operation}, \emph{recover}) \\
  & (\emph{friend}, \emph{has\_operation}, \emph{delete}) \\
  & \textbf{(\emph{friend}, \emph{has\_operation}, \emph{recover})} \\
  & \textbf{(\emph{friend}, \emph{component\_of}, \emph{WeChat})} \\
\bottomrule
\end{tabular}
\caption{An example of a query with its entities and triple candidates. The bold triples indicate the correct triples which should be selected by the query anchoring module.}
\label{tab:case1}
\end{table}

In KG construction, we first set dozens of seed entities in the target domain, and then use some patterns-based models with conventional NER models like CRF to get the final entity set. Extracting useful entities from existing knowledge bases is also a good supplement in practice. Based on these entities, we combine several models to get triple candidates. (1) We first use \emph{pattern-based bootstrapping} methods with count-based features, lexical features (e.g., term weight and POS tag) and semantic parser results to generate high-frequent triples. (2) Next, we implement some \emph{neural relation extraction} models (e.g., CNN/PCNN with attention \cite{lin2016neural}) for relation extraction. (3) We jointly consider all models with a linear transformation to rank all triple candidates. (4) Finally, we further conduct \emph{CKRL} \cite{xie2018does}\emph{ assisted by human annotation} to make sure the accuracy of KG is above 95\%.
In real-world scenarios, KG customization is labor-intensive but indispensable for high precision and interpretability in QA systems.

\subsection{Query Anchoring}

Query anchoring attempts to extract core semantics via knowledge anchors.
Simply relying on string matching or semantic parser is straightforward, while it will bring in ambiguity and noises. Moreover, semantic parsers usually perform unsatisfactory on irregular queries.
Hence, we conduct both entity and triple disambiguation models to address this issue.
For entity disambiguation, we first conduct a string matching to retrieve all possible entity candidates. For efficiency, we directly implement a forward maximum matching algorithm \cite{xiaohua2011improved} for entity disambiguation, whose accuracy is acceptable in our software scenario.

For triples, we first extract all possible connections between any two entities as triple candidates if the entity pair appears in KG. As in Table \ref{tab:case1}, there are four triple candidates of \emph{has\_operation} that reflect different core semantics. The triple disambiguation model needs to find the true purpose of the query.
We conduct an ensemble triple disambiguation model with three models.
(1) The \emph{rule-based model (RB)} considers simple syntactic rules, patterns, lexical features (e.g., token weights, POS tags, entity types), and triple-level features (e.g., relation types, entity pair distances). This model highlights valuable empirical observations and is simple and effective, where lots of general rules could be easily transferred to other domains.
(2) The \emph{knowledge reasoning model (KR)} enables some heuristic multi-hop knowledge reasoning patterns over KGs. For example, since \emph{friend} is a component of \emph{WeChat}, the target object of \emph{recover} in Fig. \ref{fig:example} is more likely to be \emph{friend} rather than \emph{WeChat}.
(3) As for the \emph{neural triple disambiguation model (NTD)}, we build a supervised model based on FastText \cite{joulin2016bag}, which takes a sentence with its target triple as the input and outputs a confidence score. The inputs are: (a) Target triple that indicates which triple candidate we focus on. (b) Position features, which show the distances from the current token to two entities in the target triple. There are two position features for each token. (c) Conflict entity features: if (\emph{$e_A$, $r$, $e_B$}) makes a triple candidate, while $e_B$ is already in the target triple (\emph{$e_C$, $r$, $e_B$}), then ($e_A$, $e_C$) is the conflict entity pair. (d) Conflict triple features: if a triple (except the target triple itself) shares any entities with that in target triple, then this triple is viewed as a conflict triple.
All features are aggregated and fed into FastText.
In practice, the knowledge reasoning model first works as a high-confident filter to remove obvious illogical results. The final triple confidence score is the weighted addition of the rule-based and neural model scores, with the weights empirically set as $0.3$ and $0.7$.

\subsection{Multi-channel Query-document Matching}

The query-document matching module takes queries and document titles with their knowledge anchors as inputs, and outputs the query-document similarity features. The input of query contains three channels, including the token sequence $W^q$, the entity sequence $E^q$ and the triple sequence $T^q$, and the same for document titles. The final similarity vector $\mathbf{s}$ is formalized as follows:
\begin{equation}
\begin{split}
\mathbf{s}=\mathrm{softmax}(\mathrm{MLP}(\mathbf{f}^{(q,d)})), \quad \mathbf{f}^{(q,d)} = \mathrm{concat} ( \mathbf{f}_w^{(q,d)},\mathbf{f}_e^{(q,d)},\mathbf{f}_t^{(q,d)} ),
\end{split}
\end{equation}
where $\mathrm{MLP}(\cdot)$ is a 2-layer perception and $\mathbf{f}^{(q,d)}$ is the aggregated query-document similarity features. $\mathbf{f}_w^{(q,d)}$, $\mathbf{f}_e^{(q,d)}$, $\mathbf{f}_t^{(q,d)}$ indicate the hidden states of query-title pairs for token, entity and triple channels respectively.
The multi-channel matching strategy jointly considers the matching degrees from different aspects with token, entity and triple.
To show the flexibility and robustness of KAQA in various situations, we learn $\mathbf{f}_w^{(q,d)}$, $\mathbf{f}_e^{(q,d)}$, $\mathbf{f}_t^{(q,d)}$  based on three representative sentence matching models including ARC-I, MatchPyramid and IWAN. It is not difficult for KAQA to use other sentence matching models.

\textbf{Architecture-I (ARC-I).}
ARC-I is a classical sentence matching model following the siamese architecture \cite{hu2014convolutional}. It first uses neural networks like CNN to get the sentence representations of both query and title separately, and then calculates their similarities. Here, $\mathbf{f}_w^{(q,d)}$ is concatenated by the final query and title representations with token sequences, and the same as $\mathbf{f}_e^{(q,d)}$ and $\mathbf{f}_t^{(q,d)}$.
\begin{equation}
\begin{split}
\mathbf{f}_w^{(q,d)} = \mathrm{Concat} (\mathrm{CNN}(\mathbf{W}^q), \mathrm{CNN}(\mathbf{Q}^d)).
\end{split}
\end{equation}

\textbf{MatchPyramid.}
Differing from ARC-I, MatchPyramid calculates the sentence similarity directly from the token-level interaction matrix \cite{pang2016text}. We use the cosine similarity to build the 2D interaction matrix. The similarity features are the hidden state after the final 2D pooling and convolution layers.
\begin{equation}
\begin{split}
\mathbf{f}_w^{(q,d)} = \mathrm{CNN}(\mathbf{M}), \quad \mathbf{M}_{ij} = \mathrm{Cosine\_sim}(\mathbf{w}_i^q, \mathbf{w}_j^d).
\end{split}
\end{equation}

\textbf{Inter-Weighted Alignment Network (IWAN).}
IWAN is an effective sentence matching model using orthogonal decomposition strategy \cite{shen2017inter}. It calculates query-document similarity based on their orthogonal and parallel components in the sentence representations. For a query, IWAN first utilizes a Bi-LSTM layer to get the hidden state $\mathbf{q}^h$ (and correspondingly $\mathbf{d}^h$ for document). Next, an query-document attention mechanism is used to generate the alignment representation of query $\mathbf{q}^a$ from all hidden embeddings in document.
The parallel and orthogonal components are formalized as follows:
\begin{equation}
\begin{split}
\mathbf{q}_i^p=\frac{\mathbf{q}_i^h \cdot \mathbf{q}_i^a}{\mathbf{q}_i^a \cdot \mathbf{q}_i^a}\mathbf{q}_i^a, \quad \mathbf{q}_i^o=\mathbf{q}_i^h-\mathbf{q}_i^p,
\end{split}
\end{equation}
in which $\mathbf{q}_i^p$ indicates the parallel component that implies the similar semantic parts of document, while $\mathbf{q}_i^o$ indicates the orthogonal component that implies the conflicts between query and document. At last, both orthogonal and parallel components of query and document are concatenated to form the final query-document similarity features as $\mathbf{f}_w^{(q,d)} = \mathrm{MLP} (\mathrm{Concat} (\mathbf{q}^p, \mathbf{q}^o, \mathbf{d}^p, \mathbf{d}^o) )$.

\subsection{Implementation Details}

The query-document matching module is considered as a classification task. We utilize a softmax layer which outputs three labels: \emph{similar}, \emph{related} and \emph{unrelated}. We use cross-entropy as our loss function, which is formalized as follows:
\begin{equation}
\begin{split}
J(\theta)=-\frac{1}{n}[\sum_{i=1}^{n}\sum_{j=1}^{3}1\{y_i=j\}\log p_i(l_j|\mathbf{s})].
\end{split}
\end{equation}
$n$ represents the number of training pair instances. $1\{y_i=j\}$ equals $1$ only if the $i$-th predicted label meets the annotated result, and otherwise equals $0$.

In this paper, we conduct KAQA concentrating on the field of software customer service.
In query anchoring module, the \emph{synonym} and \emph{hypernym/hyponym} relations are directly utilized for entity and triple normalization and generalization, while \emph{component\_of} is mainly utilized for knowledge reasoning in triple disambiguation. In query-document matching, we only consider the instances with \emph{has\_operation} as the triple part in knowledge anchors empirically, for they exactly represent the core semantics of operation. It is not difficult to consider more relation types in our multi-channel matching framework.

\section{Experiments}

\subsection{Dataset and knowledge graph}

In this paper, we construct a new dataset FAQ-SCS for evaluation, which is extracted from a real-world FAQ-based QA system in WeChat Search, since there are few large open-source FAQ datasets. In total, FAQ-SCS contains $29,134$ query-title pairs extracted from a real-world software customer service FAQ-based QA system. All query-title pairs are manually annotated with \emph{similar}, \emph{related} and \emph{unrelated} labels.
Overall, FAQ-SCS has $12,623$ \emph{similar}, $7,270$ \emph{related} and $9,241$ \emph{unrelated} labels. For evaluation, we randomly split all instances into train, valid and test set with the proportion of $8$:$1$:$1$.
We also build a knowledge graph KG-SCS in the software customer service domain for KAQA. KG-SCS contains $4,530$ entities and $4$ relations. After entity normalization via alignments with \emph{synonym} relations, there are totally $1,644$ entities and $10,055$ triples.
After query anchoring, there are $1,652$ entities and $2,877$ triples appeared in FAQ-SCS, $83.1\%$ queries and $86.7\%$ titles have at least one triple.

\subsection{Experimental Settings}

In KAQA, we implement three representative models including the siamese architecture model ARC-I \cite{hu2014convolutional}, the lexical interaction model MatchPyramid \cite{pang2016text}, and the orthogonal decomposition model IWAN \cite{shen2017inter} for sentence matching in our multi-channel matching module, with their original models considered as baselines. We do not compare with KBQA models for they are different tasks. All models share the same dimension of hidden states as $128$. In training, the batch size is set to be $512$ while learning rate is set to be $0.001$.
For ARC-I and MatchPyramid, the dimension of input embeddings is $128$. The number of filters is $256$ and the window size is $2$ in CNN encoder.
For IWAN, the dimension of input embedding is $256$. All parameters are optimized on valid set with grid search.
For fair comparisons, all models follow the same experimental settings.

\subsection{Online and Offline Query-document Matching}

\textbf{Offline Evaluation.}
We consider the evaluation as a classification task with three labels as unrelated, related or similar. We report the average accuracies across $3$ runs for all models.
From Table \ref{tab:main_experimenet} we can observe that:

\begin{table}[!hbtp]
\centering
\small
\begin{tabular}{p{5.0cm}|p{1.5cm}<{\centering}}
\toprule
Model & Accuracy  \\
\midrule
MatchPyramid \cite{pang2016text} & 0.714  \\
ARC-I \cite{hu2014convolutional} & 0.753  \\
IWAN \cite{shen2017inter} & 0.778  \\
\midrule
KAQA (MatchPyramid)  &  0.747 \\
KAQA (ARC-I) & 0.773 \\
KAQA (IWAN)  & \textbf{0.797} \\
\bottomrule
\end{tabular}
\caption{Offline evaluation on query-document matching.}
\label{tab:main_experimenet}
\end{table}

(1) The KAQA models significantly outperform all their corresponding original models on FAQ-SCS, among which KAQA (IWAN) achieves the best accuracy. It indicates that knowledge anchors and KAQA can capture core semantics precisely. We also find that pre-train models are beneficial for this task.
Moreover, KAQA performs better when there are multiple triple candidates, which implies that KAQA can distinguish useful information from noises and handle informality and ambiguity in natural language.

(2) All KAQA models with different types of sentence matching models have improvements compared to their original models. Specifically, we evaluate our KAQA framework with siamese architecture model (ARC-I), lexical interaction model (MatchPyramid) and orthogonal decomposition model (IWAN). The consistent improvements reconfirm the robustness of KAQA with different types of matching models. In real-world scenarios, KAQA can flexibly select simple or sophisticated matching models to balance both effectiveness and efficiency.

\textbf{Online Evaluation.}
To further confirm the power of the KAQA framework in real-world scenario, we further conduct an online A/B test on WeChat Search. We implement the KAQA framework with its corresponding baseline model in online evaluation. We conduct the online A/B test for $7$ days, with approximately $14$ million requests influenced by our online models. The experimental results show that KAQA achieves $1.2\%$ improvements on Click-through-rate (CTR) compared to the baseline model with the significance level $\alpha=0.01$. With the help of knowledge anchors, KAQA could have better performances in interpretability, cold start and immediate manual intervention. It has also been successfully used in other domains like medical and digital fields.

\subsection{Analysis on Query Anchoring}

In this subsection, we evaluate the effectiveness of different triple disambiguation models.
We construct a new triple disambiguation dataset for query anchoring evaluation. Specifically, we randomly sample queries from a real-world software customer service system. To make this task more challenging, we only select the complicated queries which have at least two triple candidates with \emph{has\_operation} relation before triple disambiguation. At last, we sample $9,740$ queries with $20,267$ triples. After manually annotation, there are $10,437$ correct triples that represent the core semantics, while the rest $9,830$ triples are incorrect. We randomly select $1,877$ queries for evaluation.

There are mainly three triple disambiguation components. We use RB to indicate the basic rule-based model, KR to indicate the knowledge reasoning model, and NTD to indicate the neural triple disambiguation model. We conduct three combinations to show the contributions of different models, using Accuracy and AUC as our evaluation metrics. In Table \ref{tab:query_anchoring}, we can find that:

\begin{table}[!hbtp]
\centering
\small
\begin{tabular}{p{4.0cm}|p{1.5cm}<{\centering}|p{1.5cm}<{\centering}}
\toprule
Model & Accuracy & AUC \\
\midrule
KAQA (RB) & 0.588 & 0.646 \\
KAQA (RB+KR) & 0.619 & 0.679 \\
KAQA (RB+KR+NTD) & \textbf{0.876} & \textbf{0.917} \\
\bottomrule
\end{tabular}
\caption{Results of triple disambiguation.}
\label{tab:query_anchoring}
\end{table}

(1) The ensemble model RB+KR+NTD that combines all three disambiguation components achieves the best performances on both accuracy and AUC. User queries in FAQ-based QA usually struggle with abbreviations, informal representations and domain-specific conditions. The results reconfirm that our triple disambiguation model is capable of capturing user intention precisely, even with the complicated queries containing multiple triple candidates. We will give detailed analysis on such complicated queries in case study.

(2) The neural triple disambiguation component brings in huge improvements compared to rule-based and knowledge reasoning models. It indicates that the supervised information and the generalization ability introduced by neural models are essential in triple disambiguation. Moreover, RB+KR model significantly outperforms RB model, which verifies that knowledge-based filters work well.

\subsection{Ablation Tests on Knowledge Anchors}

In this subsection, we attempt to verify that all components of KAQA are effective in our task. We set two different settings, the first removes triples in knowledge anchors, while the second removes entities. We report the accuracies of these two settings on KAQA (ARC-I) in Table \ref{tab:effective_mechanisms}. We find that both settings have consistent improvements over the original models, which also implies that the entities and triples are useful for matching. Moreover, triples seem to play a more essential role in knowledge anchors.

\begin{table}[!hbtp]
\centering
\small
\begin{tabular}{p{4.0cm}|p{1.5cm}<{\centering}}
\toprule
Model & Accuracy  \\
\midrule
ARC-I & 0.753 \\
KAQA (ARC-I) (entity) & 0.762  \\
KAQA (ARC-I) (triple) & 0.766  \\
KAQA (ARC-I) (all) & \textbf{0.773}  \\
\bottomrule
\end{tabular}
\caption{Results of different knowledge anchors.}
\label{tab:effective_mechanisms}
\end{table}

\subsection{Case Study}

\begin{table*}[!htbp]
\center
\small
\begin{tabular}{p{4.2cm}|p{4.2cm}|p{1.0cm}<{\centering}|p{1.0cm}<{\centering}|p{1.0cm}<{\centering}}
 \toprule
  Query & Title & ARC-I & KAQA & Label\\
 \midrule
 How to delete WeChat's chat logs. & Can WeChat recover chat logs that have been deleted? & 2 & 0 & 0\\
 (\emph{chat log}, \emph{OP}, \emph{delete}) & (\emph{chat log}, \emph{OP}, \emph{recover}) & & &\\
 \midrule
 How can I not add pictures (when sending messages) in Moments? & In Moments, can I only share textual messages without attaching figures? & 0 & 2 & 2 \\
 (\emph{picture}, \emph{OP}, \emph{(not) add}) & (\emph{figure}, \emph{OP}, \emph{(not) attach}) & & &\\
 \midrule
 How to log in WeChat with new account? & Can I log in WeChat with two accounts simultaneously
? & 2 & 2 & 1 \\
 (\emph{(new) account}, \emph{OP}, \emph{log in}) & (\emph{(two) account}, \emph{OP}, \emph{log in}) & & &\\
 \midrule
 What should I do to set up administrators in the group? & How to change the administrator in my chatting group? & 0 & 0 & 2 \\
 (\emph{administrator}, \emph{OP}, \emph{set up}) & (\emph{administrator}, \emph{OP}, \emph{change}) & & &\\
 \bottomrule
\end{tabular}
\caption{\label{tab:case_study} Examples of query-title matching with triples and labels. Label 0/1/2 indicates unrelated/related/similar. We only show the triples that indicate core semantics.}
\end{table*}

In Table \ref{tab:case_study}, we give some representative examples to show the pros and cons of using knowledge anchors.
In the first case, KAQA successfully finds the correct knowledge anchor \emph{(chat log, OP, recover)} in title via the triple disambiguation model, avoiding confusions caused by the candidate operation \emph{delete}. While the original ARC-I model makes a mistake by only judging from tokens.
In the second case, there is a semantic ellipsis (send messages) in user query that confuses ARC-I, which usually occurs in QA systems. However, KAQA successfully captures the core semantics (\emph{picture}, \emph{OP}, \emph{(not) add}) to get the right prediction. The \emph{synonym} relation also helps the alignment between ``figure" and ``picture". However, KAQA also has limitations.
In the third case, knowledge anchors merely concentrate on the core semantic operation \emph{log in WeChat account}, paying less attention to the differences between ``new" and ``two". Therefore, KAQA gives a wrong prediction of \emph{similar}. A more complete KG is needed.
In the last case, KAQA does extract the correct knowledge anchors. However, although \emph{set up} and \emph{change} have different meanings, \emph{set up}/\emph{change} \emph{administrator} should indicate the same operation in such scenario. Consider the \emph{synonym} and \emph{hypernym/hyponym} relationships between triples will partially solve this issue.

\section{Conclusion and Future Work}

In this paper, we propose a novel KAQA framework for real-world FAQ systems. We consider entities and triples in texts as knowledge anchors to precisely capture core semantics for NLU and matching. KAQA is effective for real-world FAQ systems that pursue high precision, better interpretability with faster and more controllable human intervention, which could be rapidly adapted to other domains.
Experimental results confirm the effectiveness and robustness of KAQA.

We will explore the following research directions in future: (1) We will consider more sophisticated and general methods to fuse knowledge anchors into the multi-channel matching module. (2) We will explore the relatedness between entity and triple to better modeling knowledge anchor similarities.

\bibliography{reference}
\bibliographystyle{splncs04}

\end{document}